\documentclass[letterpaper]{article}
\usepackage{aaai}
\usepackage{times}
\usepackage{helvet}
\usepackage{courier}

\usepackage{graphicx}
\usepackage{amssymb}
\usepackage{amsmath}
\usepackage{multicol}
\usepackage{color, colortbl}
\usepackage[table,xcdraw]{xcolor}
\usepackage{booktabs}
\usepackage{comment}
\usepackage{placeins}
\usepackage{bbding}

\usepackage{enumitem}

\begin{document}
% The file aaai.sty is the style file for AAAI Press 
% proceedings, working notes, and technical reports.
%
\title{NAS-based Recursive Stage Partial Network (RSPNet) for Light-Weight Semantic Segmentation}
\author{Anonymous Authors\\
}
\maketitle

%%%%%%%%%%%%%%%%%%%%%%%%%%%%%%%%%%%%%%
%%%%%%%%%%%%%%%%%%%%%%%%%%%%%%%%%%%%%%
\begin{abstract}
 Current NAS-based semantic segmentation methods focus on accuracy improvements rather than light weight design.  In this paper, we propose a two-stage framework to design our NAS-based RSPNet model for light-weight semantic segmentation. The first architecture search determines the inner cell structure, and the second architecture search considers exponentially growing paths to finalize the outer structure of the network.  It was shown in the literature that the fusion of high- and low-resolution feature maps produces stronger representations. To find the expected macro structure without manual design, we adopt a new path-attention mechanism to efficiently search for suitable paths to fuse useful information for better segmentation. Our search for repeatable micro-structures from cells leads to a superior network architecture in semantic segmentation. In addition, we propose an RSP (recursive Stage Partial) architecture to search a light-weight design for NAS-based semantic segmentation.  The proposed architecture is very efficient, simple, and effective that both the macro- and micro- structure searches can be completed in five days of computation on two V100 GPUs. The light-weight NAS architecture with only 1/4 parameter size of SoTA architectures can achieve SoTA performance on semantic segmentation on the Cityscapes dataset without using any backbones.
\end{abstract}
%%%%%%%%%%%%%%%%%%%%%%%%%%%%%%%%%%%%%%
%%%%%%%%%%%%%%%%%%%%%%%%%%%%%%%%%%%%%%

%%%%%%%%%%%%%%%%%%%%%%%%%%%%%%%%%%%%%%
%%%%%%%%%%%%%%%%%%%%%%%%%%%%%%%%%%%%%%
%%%%%%%%%%%%%%%%%%%%%%%%%%%%%%%%%%%%%%
\section{Introduction}

{\em Network Architecture Search} (NAS)~\cite{NAS:Survey:JMLR2019} is a computational approach for automating the optimization of the neural architecture design. As deep learning has been widely used for medical image segmentation, the most common deep networks used in practice are still designed manually.  In this work, we focus on applying NAS for {\em medical image segmentation} as the targeted application. To optimize the NAS that looks for the best architecture for image segmentation,
the search task can be decomposed into three parts: 
{\em (i)} a supernet to generate all possible architecture candidates,
{\em (ii)} a global search of {\em neural architecture paths} from the supernet, and
{\em (iii)} a local search of the {\em cell architectures}, namely operations including the conv/deconv kernels and the pooling parameters. 
The NAS space to explore is exponentially large {\em w.r.t.} the number of generated candidates, the paths between nodes, the number of depths, and the available cell operations to choose from. The computational burden of NAS for image segmentation is much higher than other tasks such as image classification, so each architecture verification step takes longer to complete. As a result, there exist fewer NAS methods that work successfully for image segmentation.  In addition, none is designed for light-weight semantic segmentation, which is very important for AV(Automobile Vehicle)-related applications.

%--------------------------------------
\begin{figure*}[t]
\centerline{
  \includegraphics[height=0.45\linewidth]{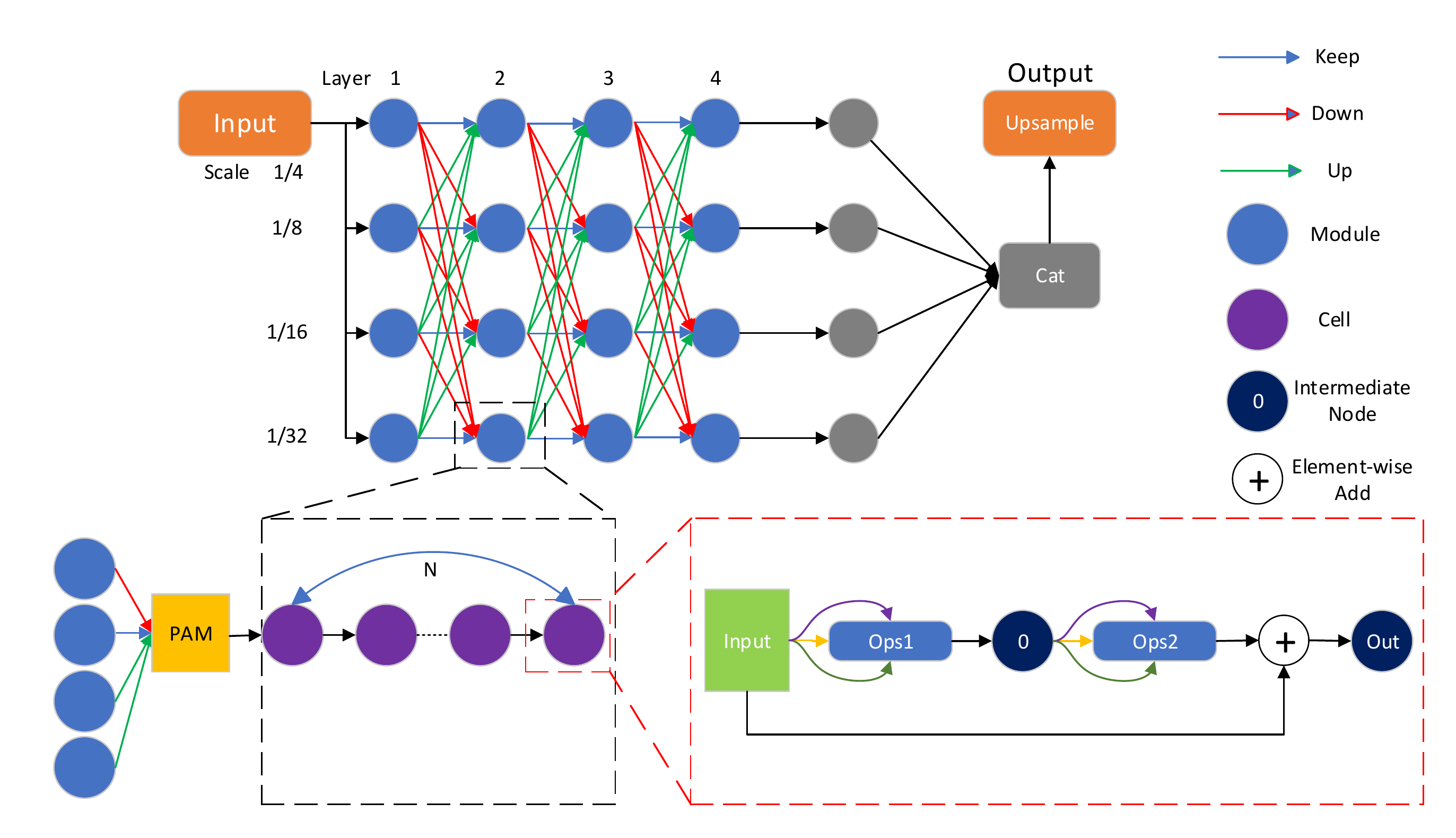}
  \vspace{-0.3cm}
}  
\caption{
{\em Top:} Our fully connected model with Path-Attention Module(PAM) to search macro structure. {\em Bottom Left:} Before entering the next layer, feature-maps generated by previous layer pass through the Path-Attention Module(PAM) and choose the best one to enter the cell. {\em Bottom Right:} Because of the efficiency and parameter size, we only use one cell at each layer when we search cell using DARTS. During training, we stack four cells we searched in search process.
} 
\label{fig:overview}
\vspace{-0.4cm}
\end{figure*}
%--------------------------------------

The main challenge of NAS is on how to deal with the exponentially large search space when exploring and evaluating neural architectures.
We tackle this problem based on a formulation regarding {\em what needs to be considered in priority and how to effectively reduce search complexity}. 
Most segmentation network designs~\cite{UNet:MICCAI2015,GridNet:BMVC2017,NAS:UNet:IEEEAccess2019,AutoDeepLab:CVPR2019} use U-Nets to achieve better accuracies in image segmentation.  For example, AutoDeepLab~\cite{AutoDeepLab:CVPR2019} designs a level-based U-Net as the supernet whose search space grows exponentially according to its level number $L$ and depth parameter $D$. Joining the search for network-level and cell-level architectures creates huge challenges and inefficiency in determining the best architecture. To avoid exponential growth in the cell search space, only one path in AutoDeepLab is selected and sent to the next node.  This limit is unreasonable since more input to the next node can generate richer features for image segmentation. 

A ``repeatable'' concept is adopted in this paper to construct our model. Similar to repeatable cell architecture design, our model contains repeated units that share the same structure. The proposed model architecture for image segmentation is shown in Fig.~\ref{fig:overview}. It is based on differential learning and is as efficient as the DARTS (Differentiable ARchiTecture Searching) method~\cite{DARTS:ICLR2019}, compared to the other NAS methods based on RL and EV.  In addition, we modify the concept of CSPNet~\cite{CSPNet:CVPRW2020} to recursively use only half of the channels to pass through the cell we searched for. The RSPNet (Recursive Stage Partial Network) makes our search procedure much more efficient and results in a light weight architecture for semantic segmentation.  The proposed architecture is simple, efficient, and effective in image segmentation.  Both the macro- and micro- structure searches can be completed in two days of computation on two V100 GPUs. The light-weight NAS network with only 1/4 parameter size of SoTA architectures achieves state-of-the-art performance on the Cityscapes datasets without any backbones. Main contributions of this paper are summarized in the following:
\begin{quote}
\begin{itemize}
\item We propose a two-stage search method to decrease memory usage and speed up search time. 

\item We designed a cell-based architecture that can construct a complex model by stacking the cell we searched for.

\item RSPNet makes our search procedure much more efficient and results in our light-weight architecture for semantic segmentation.   

\item The proposed PAM selects the paths and fuses more inputs than 
Auto-Deeplab~\cite{AutoDeepLab:CVPR2019} and HiNAS~\cite{HiNAS:CVPR2020}
to generate richer features for better image segmentation.

\item Without using any backbone, our architecture outperforms SoTA methods with improved accuracy and efficiency on the Cityscapes~\cite{Cityscapes:CVPR2016} dataset.
\end{itemize}
\end{quote}

%%%%%%%%%%%%%%%%%%%%%%%%%%%%%%%%%%%%%%
%%%%%%%%%%%%%%%%%%%%%%%%%%%%%%%%%%%%%%
%%%%%%%%%%%%%%%%%%%%%%%%%%%%%%%%%%%%%%
\section{Related Works}

Mainstream NAS algorithms generally consist of three basic steps: {\em supernet generation}, {\em architecture search}, and {\em network cell parameter optimization}. 
Approaches for architecture search can be organized into three categories~\cite{NAS:Survey:JMLR2019}: {\em reinforcement learning (RL)} based, {\em evolution (EV)} based, and {\em gradient} based.  In what follows, the details of each method are discussed.
%differential architecture search (DARTS) based~\cite{DARTS:ICLR2019}.

{\bf RL-based methods} \cite{Baker:NAS:RL:ICLR2017,Zoph:NAS:RL:ICLR2017} use a controller to sample neural network architectures (NNAs) to learn a reward function to generate better architectures from exploration and exploitation.
 Although an RL-based NAS approach can construct a stable architecture for evolution, it needs a huge number of tries to get a positive reward for updating architectures and thus is very computationally expensive.  For example, in \cite{BlockQNN:CVPR2018}, a cell-based Q-learning approach is evaluated on ImageNet, and the search takes up to 9 GPU-days to run.

{\bf Evolution-based methods}~\cite{Largescale:Img:Class:ICML2017,AmoebaNet:AAAI2019,Hierarchical:NAS:ICLR2018,Efficient:NAS:ICLR2019} perform evolution operators ({\em e.g.} crossover and mutation) based on the genetic algorithm (GA) to continuously adjust NNAs and improve their qualities across generations. These methods suffer from high computational cost in optimizing the model generator when validating the accuracy of each candidate architecture. Compared to the methods, which rely on optimization over discrete search spaces, gradient-based methods can optimize much faster via search in continuous spaces.

{\bf Gradient-based methods:} The Differentiable ARchiTecture Search (DARTS)~\cite{DARTS:ICLR2019} signiﬁcantly improves search efﬁciency by computing a convex combination of a set of operations where the best architecture can be optimized by gradient descent algorithms. In DARTS, a {\em supernet} is constructed by placing a mixture of candidate operations on each edge rather than applying a single operation to a node. An {\em attention mechanism} on the connections is adopted to remove weak connections, such that all supernet weights can be efficiently optimized jointly with a continuous relaxation of the search space via gradient descent. The best architecture is found efficiently by restricting the search space to the subgraphs of the supernet.
However, this acceleration comes with a penalty when limiting the search variety, resulting in architectures with inferior accuracy when compared with RL- or EV-based methods.

{\bf Nework parameter sharing methods}~\cite{Efficient:NAS:ICML2018,CARS:NAS:CVPR2020,Trans:Arch:CVPR2018,NAS:Net:Transform:AAAI2018,SMASH:NAS:ICLR2018,N2N:ICLR2018} use a weight-sharing concept to reduce gradient updates on network parameters.  For example, in \cite{N2N:ICLR2018}, the knowledge of well-trained architectures is compressed and transferred through network compression operations to improve the efficiency and effectiveness of model learning.

{\bf NAS for image segmentation} is less investigated due to large memory demands and model validation costs.
In \cite{NAS:UNet:IEEEAccess2019}, a U-like backbone is used to search down-sampling and up-sampling cells of repeatable structures for medical image segmentation.
The NAS-UNet~\cite{NAS:UNet:IEEEAccess2019} is a U-like architecture, where the micro structure is automatically adjusted by a differential search for a repeatable cell structure.
In AutoDeepLab~\cite{AutoDeepLab:CVPR2019}, a two-level hierarchical architecture search space is used to find the best architecture. Similar to \cite{Trans:Arch:CVPR2018}, the first search is applied for the inner cell structure search, and the second search considers exponential-many paths to determine the outer structure of the network. Significant results are achieved, but with 3 days of GPU computation.

%
%{\bf Fully-Connected NAS:} The method of feature fusing in \cite{AutoDeepLab:CVPR2019,HiNAS:CVPR2020} takes only one previous layers' outputs as inputs. Different from AutoDeeplab~\cite{AutoDeepLab:CVPR2019} and HiNAS~\cite{HiNAS:CVPR2020}, 
%our feature fusion method takes all previous layers’ outputs as inputs to next cell we search. Our method can fuse much more information to generate better segmentation results. 
%

{\bf Path-Attention NAS:} The method of path searching mentioned in \cite{AutoDeepLab:CVPR2019,HiNAS:CVPR2020} takes only three previous layers' outputs as inputs and uses the Viterbi
decoding algorithm to select the path with the maximum
probability as the final result. Different from AutoDeeplab~\cite{AutoDeepLab:CVPR2019} and HiNAS~\cite{HiNAS:CVPR2020}, 
our path selection method takes all previous layers’ outputs and current layer’s output as inputs 
 and uses the DARTS algorithm to choose the best two layers as inputs to next cells we search.   
 Our path-attention method can fuse much more information from more inputs to generate better segmentation results.
 
%------------------------------
\begin{figure}[t]
\centerline{
  \includegraphics[width=1.1\linewidth]{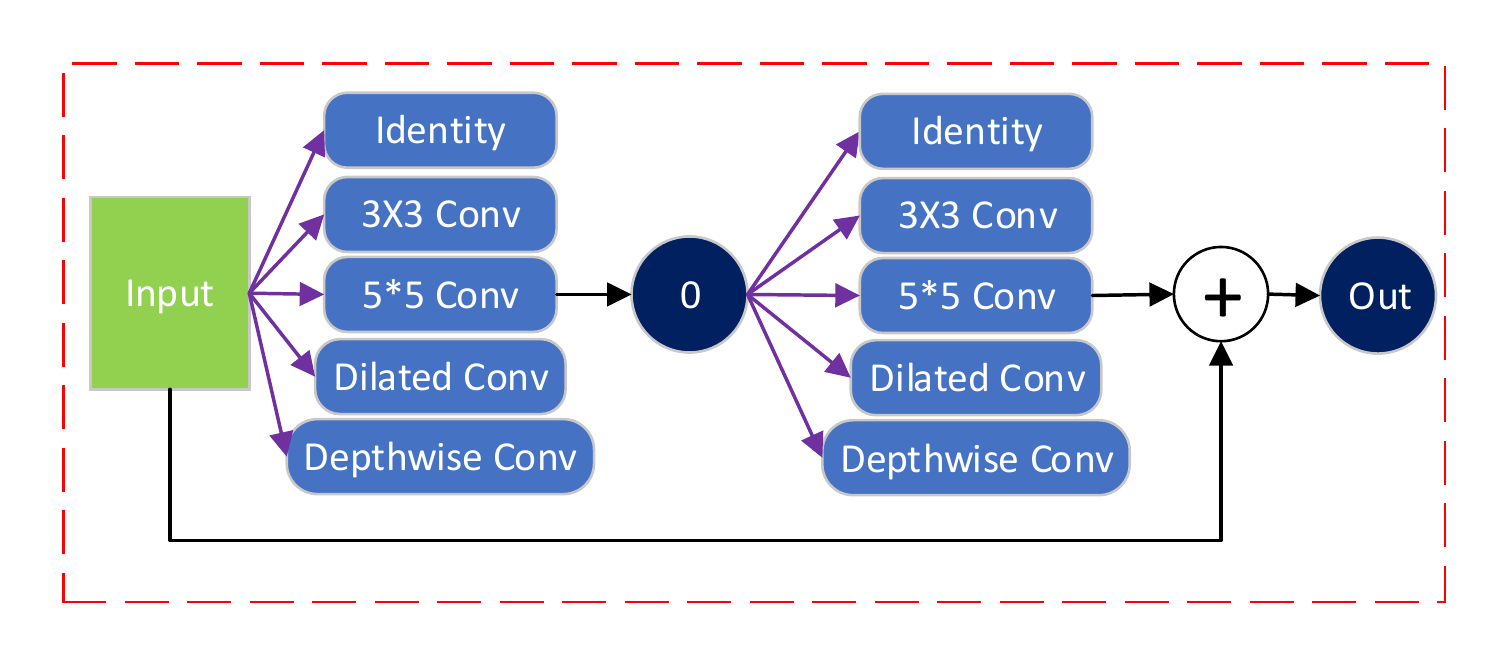}
  \vspace{-0.2cm}
}  
\caption{
The Cell architecture we propose.
} 
\label{fig:N:cell}
\vspace{-0.2cm}
\end{figure}
%------------------------------

%%%%%%%%%%%%%%%%%%%%%%%%%%%%%%%%%%%%%%
%%%%%%%%%%%%%%%%%%%%%%%%%%%%%%%%%%%%%%
%%%%%%%%%%%%%%%%%%%%%%%%%%%%%%%%%%%%%%
\section{Background} \label{sec:background}

%--------------------------------------
\begin{figure}[t]
\centerline{
  \includegraphics[width=\linewidth]{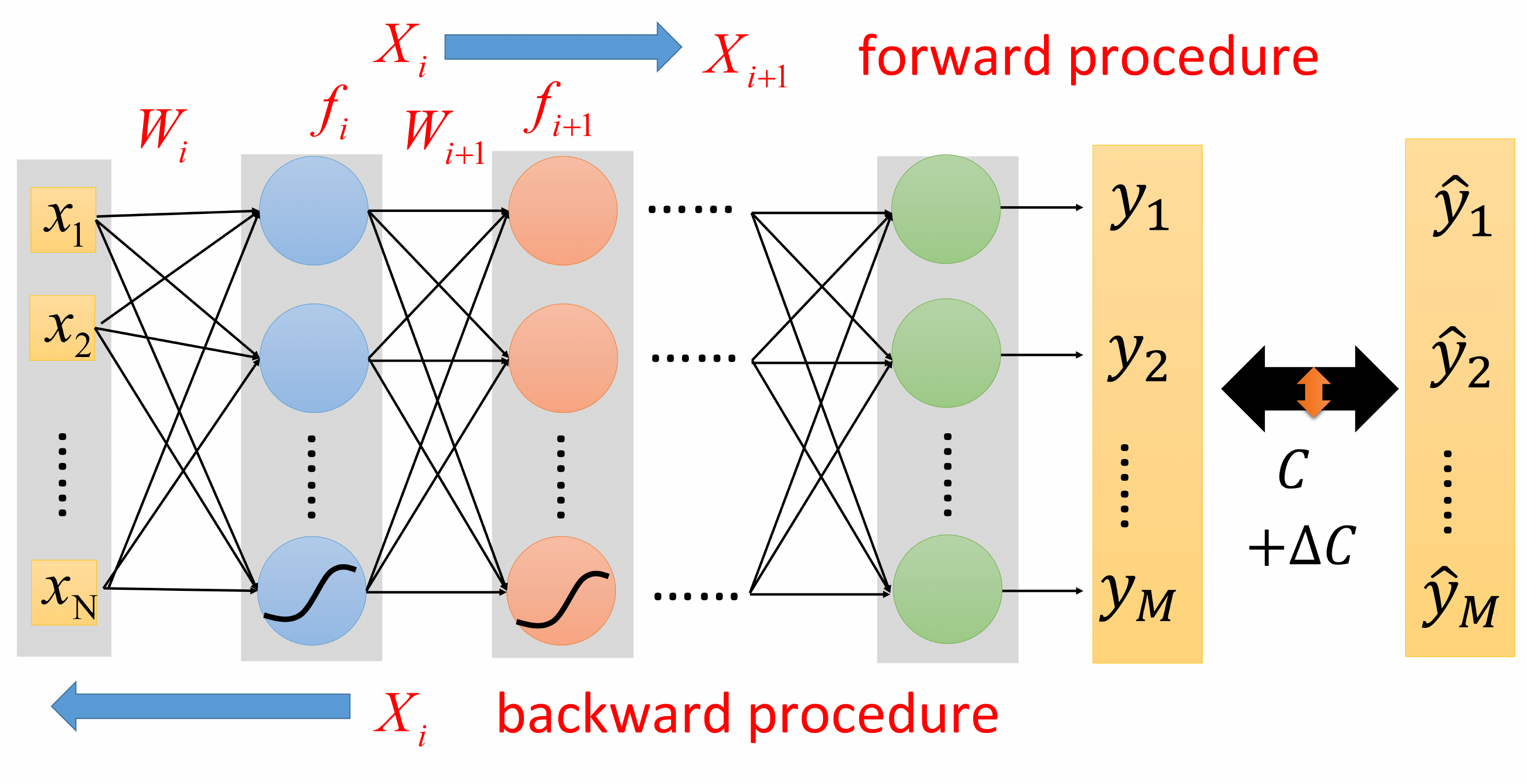}
%\vspace{-0.2cm}
}
\caption{ 
Network forward and backward procedures, where $X_i$ denotes the output of the $i$th layer with the weight matrix $W_{i+1}$ and the output function $f_{i+1}$. 
}
\label{fig:backprogation}
\vspace{-0.1cm}
\end{figure}
%---------------------------------

%--------------------------------------
\begin{figure}[t]
\centerline{
  \includegraphics[width=\linewidth]{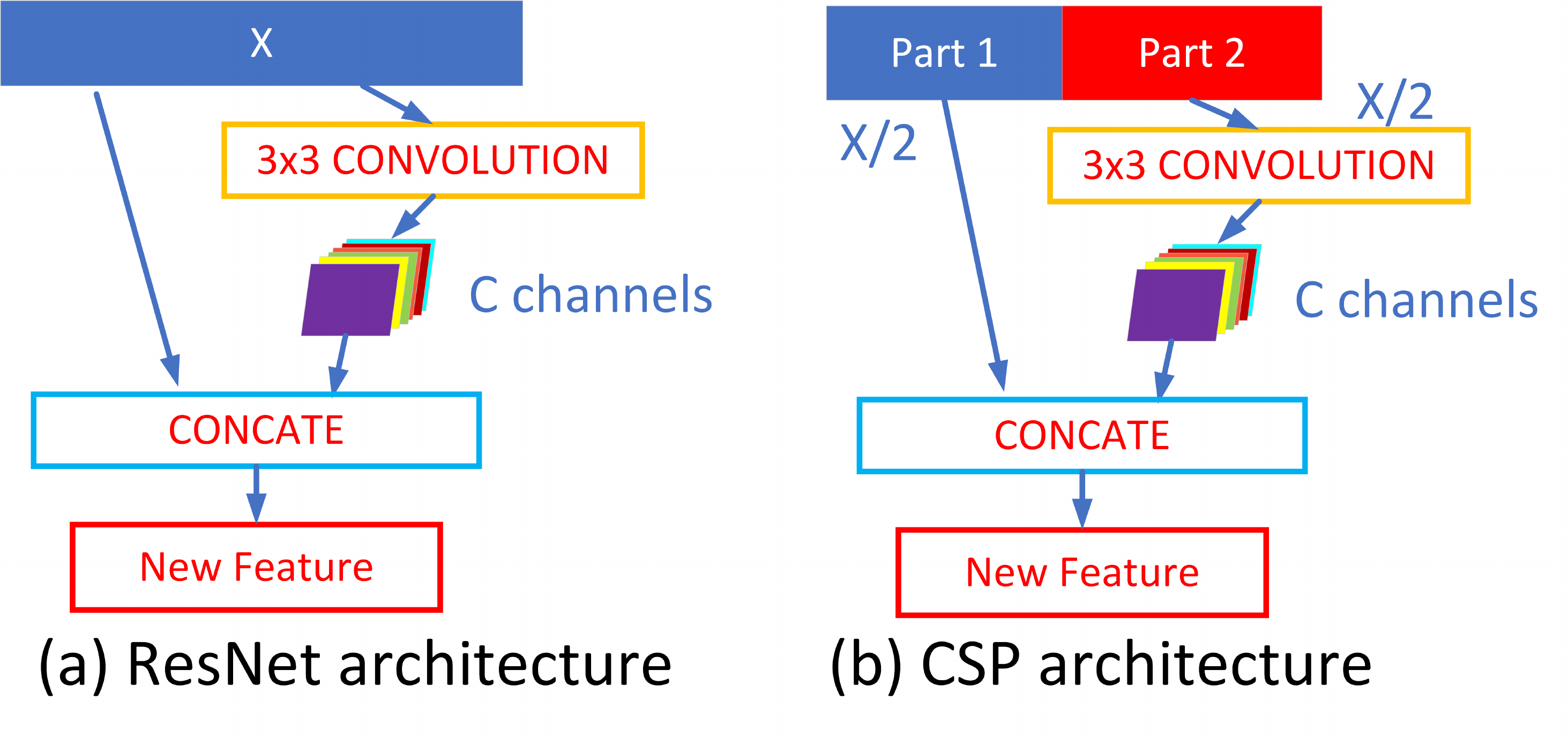}
%\vspace{-0.2cm}
}
\caption{ 
Comparisons between ResNet and CSPNet.  (a) ResNet.  (b) CSPNet. The CSPNet  separates the  feature  map  into two  parts, where one is sent to a convolution block and another is directly combined via a skip connection.
}
\label{fig:ResNetCSPNet}
\vspace{-0.4cm}
\end{figure}
%------------------------------

This section first addresses the vanishing gradient problem, which often occurs when training a neural network from a backward procedure.  Then, we discuss the relation between the vanishing gradient problem and CSPNet~\cite{CSPNet:CVPRW2020}.  Then, we modify the concept of the CSPNet 
as RSPNet that recursively uses only half of the channels to pass through the cell we searched for.  RSPNet makes our search procedure more efficient in finding the desired light-weight architecture for semantic segmentation.   

Fig.~\ref{fig:backprogation} shows the forward and backward procedures of a neural network. 
 Let $X_i$ denote the output of the $i$th layer with the weight matrix $W_{i+1}$ and the output function $f_{i+1}$.   The relation between  $X_i$ and  $X_{i+1}$ can be written as follows:
\begin{equation}
\label{eq:LayerRelation}
    X_{i+1} = f_{i+1}\left (X_i,W_{i+1} \right ).
\end{equation}
Let $C$ be a cost function. For the $(L-1)$th layer, according to the chain rule, the gradient of its weight matrix $W_{L-1}$ can be written as 
\begin{equation}
\label{eq:GradientforLayer1}
\frac{\partial }{{\partial {W_{L - 1}}}}C = \frac{{\partial C}}{{\partial {X_L}}} \frac{{\partial {X_L}}}{{\partial {X_{L-1}}}}\frac{{\partial {X_{L - 1}}}}{{\partial {W_{L - 1}}}}.
\end{equation}
Based on Eq.(\ref{eq:LayerRelation}), we have $X_L = f_L$.  Then, Eq.(\ref{eq:GradientforLayer1}) can be rewritten as
\begin{equation}
\label{eq:GradientforLayer1_2}
\frac{\partial }{{\partial {W_{L - 1}}}}C = \frac{{\partial C}}{{\partial {X_L}}} \frac{{\partial {f_L}}}{{\partial {X_{L-1}}}}\frac{{\partial {f_{L - 1}}}}{{\partial {W_{L - 1}}}}.
\end{equation}
For the $(L-2)$th layer, according to Eq.(\ref{eq:GradientforLayer1_2}) and the chain rule, the gradient of $W_{L-2}$ to $C$ can be updated as follows. 
\begin{equation}
\label{eq:GradientforLayer2}
\frac{\partial }{{\partial {W_{L - 2}}}}C = \frac{{\partial C}}{{\partial {X_L}}}\left( {\prod\limits_{k = 1}^2 {\frac{{\partial {f_{L-k+1}}}}{{\partial {X_{L - k}}}}} } \right)\frac{{\partial {f_{L - 2}}}}{{\partial {W_{L - 2}}}}.
\end{equation}
For the $(L-i)$th layer, based on the form of Eq.(\ref{eq:GradientforLayer2}), the gradient of $W_{L-i}$ to $C$ can be recursively updated by a backpropagation method, $i.e.$, 
\begin{equation}
\label{eq:BackPropagation}
\frac{\partial }{{\partial {W_{L - i}}}}C = \frac{{\partial C}}{{\partial {X_L}}}\left( {\prod\limits_{k = 1}^i {\frac{{\partial {f_{L-k+1}}}}{{\partial {X_{L - k}}}}} } \right)\frac{{\partial {f_{L - i}}}}{{\partial {W_{L - i}}}}.
\end{equation}
When the activation function used in $f_i$ is a Sigmoid function, the recursive term in Eq.(\ref{eq:BackPropagation}) will close to zero and result in the problem of vanishing gradient; that is,
\begin{equation}
\label{eq:BackPropagationInfinite}
\prod\limits_{k = 1}^i {\frac{{\partial {f_{L-k+1}}}}{{\partial {X_{L - k}}}}}  \to 0.
\end{equation}
To solve this problem in Eq.(\ref{eq:BackPropagationInfinite}), one solution is to change the activation function to the ReLU function. Another solution proposed in ResNet ~\cite{ResNet:CVPR2016} explicitly preserves information through additive identity transformation, $i.e.$, skip connection.  With the skip connection,  Eq.(\ref{eq:LayerRelation}) can be rewritten as 
\begin{equation}
\label{eq:ResidualLayerRelation}
    X_{i+1} = f_{i+1}\left ( X_i,W_{i+1} \right ) + X_i.
\end{equation}
Fig. \ref{fig:ResNetCSPNet}(a) shows the architecture of ResNet~\cite{ResNet:CVPR2016}. Based on Eq.(\ref{eq:ResidualLayerRelation}), Eq.(\ref{eq:BackPropagation}) can be rewritten as
\begin{equation}
\label{eq:ResidualFormulation}
\frac{\partial }{{\partial {W_{L - i}}}}C = \frac{{\partial C}}{{\partial {X_L}}}\left( {\prod\limits_{k = 1}^i {\frac{{\partial {f_{L-k+1}}}}{{\partial {X_{L - k}}}}} } + 1 \right)\frac{{\partial {f_{L - i}}}}{{\partial {W_{L - i}}}}.
\end{equation}
Based on Eq.(\ref{eq:ResidualFormulation}), the recursive term in  Eq.(\ref{eq:BackPropagation}) will not converge to zero, and thus the problem of vanishing gradient can be solved.  Unlike ResNet~\cite{ResNet:CVPR2016}, CSPNet~\cite{CSPNet:CVPRW2020} adopts a split-transform-merge strategy to separate the feature map of the base layer into two parts; one part will go through a convolution block and a transition layer; the other part is then combined with the transmitted feature map to the next stage. Then, Eq.(\ref{eq:ResidualLayerRelation}) can be modified as 
\begin{equation}
\label{eq:CSPLayerRelation}
    X_{i+1} = f_{i+1}\left ( {1 \over 2} X_i,W_{i+1} \right ) + {1 \over 2} X_i.
\end{equation}
It is another way to make the recursive term in Eq.(\ref{eq:BackPropagation}) not converge to zero, and thus the problem of vanishing gradient is solved.  Eq.(\ref{eq:CSPLayerRelation}) inspires the design of our RSPNet. The architecture of this CSPNet is shown in Fig.\ref{fig:ResNetCSPNet}(b). Due to the split-transform-merge strategy, rich features can be extracted and thus enhance both the efficiency and accuracy of object detection.  However, CSPNet separates the feature map into two parts only on the base layer.  We can apply the split-transform-merge strategy recursively not only on the base layer but also to other layers. The recursively mixed two-path way can preserve feature reusing characteristics, but at the same time prevents an excessive amount of duplicate gradient information by truncating the gradient ﬂow. This is named as RSPNet (Recursive Stage Partial Network).

%%%%%%%%%%%%%%%%%%%%%%%%%%%%%%%%%%%%%%
%%%%%%%%%%%%%%%%%%%%%%%%%%%%%%%%%%%%%%
%%%%%%%%%%%%%%%%%%%%%%%%%%%%%%%%%%%%%%
\section{Methods}

A new concept is proposed in this paper to use two-stage method from which the best architecture can be searched via NAS. Fig.~\ref{fig:overview} shows the architecture created in our method.  At first, the cell-structure will be first searched with light-weight structure by NAS (see {\em Bottom Right}).  Then, we fix the cell-structure and repeat it four times to construct our architecture. In what follows, the details of our method are described.   

%%%%%%%%%%%%%%%%%%%%%%%%%%%%%%%%%%%%%%
%%%%%%%%%%%%%%%%%%%%%%%%%%%%%%%%%%%%%%
\subsection{Cell Architecture Search}
\label{sec:cell:search}

For cell structure search, only one type of cell is used: the {\bf Normal Cell (Norm-Cell)} as in Fig.~\ref{fig:N:cell}. We use two intermediate nodes to construct each cell structure. Each cell keeps the number of channels and neither reduces the spatial dimension from $W \times H$ to $\frac{W}{2} \times \frac{H}{2} $ nor enlarges the spatial dimension from $W \times H$ to $2W \times 2H$.

Referring to Fig.~\ref{fig:overview} ({\em bottom left}) and Fig.~\ref{fig:N:cell}, for each Norm-Cell with one input, feature maps derived by previous layer with different shapes are reshaped by convolution.
The input maps are then fed into subsequent operations after adding together, which are searched by DARTS to generate a feature map with the same size $C \times {H} \times {W}$.  

The set of operation types, consists of the following 5 operators, all prevalent in modern CNNs:

\begin{multicols}{2}
\begin{itemize}

\item $3 \times 3$ Convolution  
\item $5 \times 5$ Convolution
\item Dilated Convolution  
\item Depthwise Convolution
\item Identity

\end{itemize}
\end{multicols}

 We list all primitive operations used for cell search that appear in Fig.~\ref{fig:N:cell}. 
For all edges between two intermediate nodes, their weights are calculated by applying softmax to obtain real-valued architecture parameters.

%%%%%%%%%%%%%%%%%%%%%%%%%%%%%%%%%%%%%%
%%%%%%%%%%%%%%%%%%%%%%%%%%%%%%%%%%%%%%
\subsection{Two Stage Search}
\label{sec:method:search}

The main challenge of NAS is on how to deal with the exponentially large search space when exploring and evaluating neural architectures.
We use a two-stage method to search and train our model. In the first stage, we adopt the lightweight architecture by using fewer cells to search. In Fig.~\ref{fig:overview} ({\em bottom left}), we only use one cell in each layer.
After the search was completed, we stack the cells searched using DARTS~\cite{DARTS:ICLR2019} four times to construct our model. Our experiment result shows that the method can not only get better efficiency during the search process, but also outperform SoTA methods with improved accuracy. 

\begin{figure*}[t]
\centerline{
  {\footnotesize (a)}
  \includegraphics[height=0.28\linewidth]{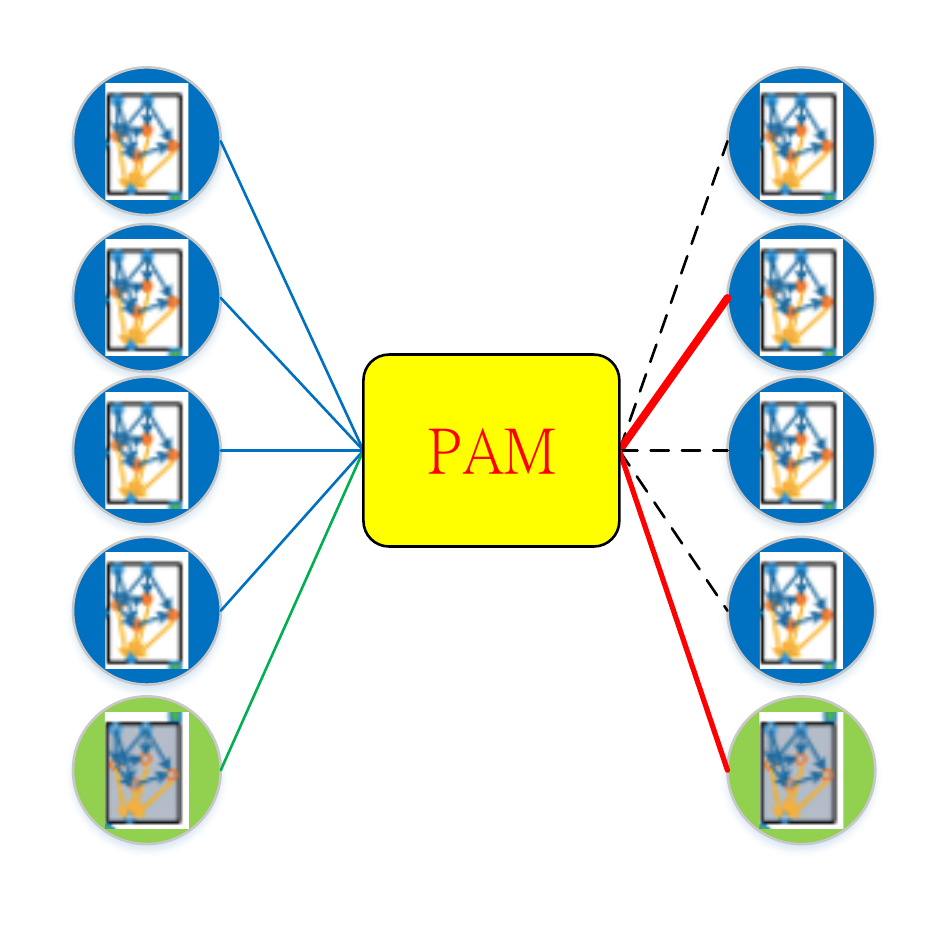}
 {\footnotesize (b)}
  \includegraphics[height=0.28\linewidth]{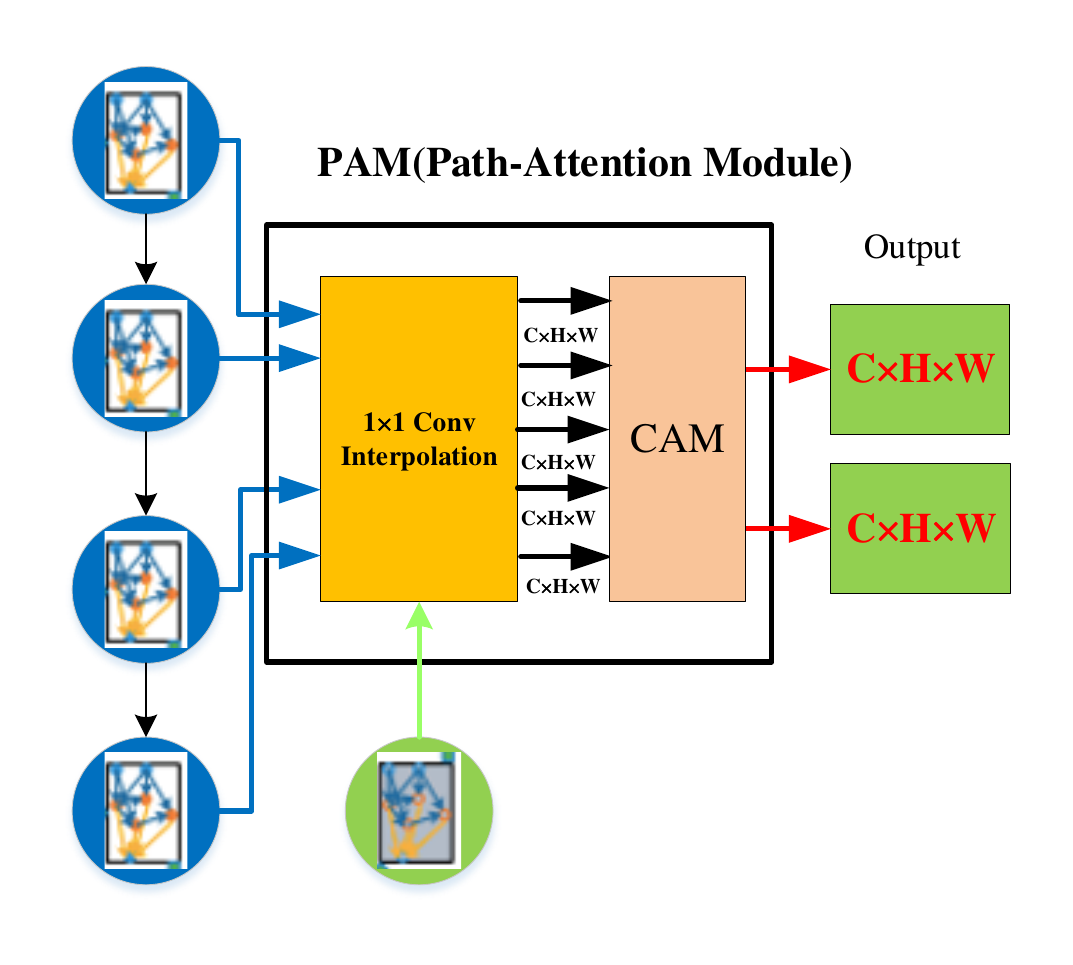}
} 
\caption{
Details of the proposed {\bf Path-Attention NAS} components. 
(a) The outputs of cells are concatenated as inputs to the {\em Path Attention Module} (PAM). Its output channels are divided into the specified number of outputs (red color)  and pass through the subsequent cell.
%\vspace{-0.4cm}
(b) Path-Attention Module is mainly to choose the number of feature maps we want, then it pass through the next cell.
} 
\label{fig:PAM}
\vspace{-0.2cm}
\end{figure*}

%--------------------------------------
\begin{figure}[t]
\centerline{

  \includegraphics[height=0.45\linewidth]{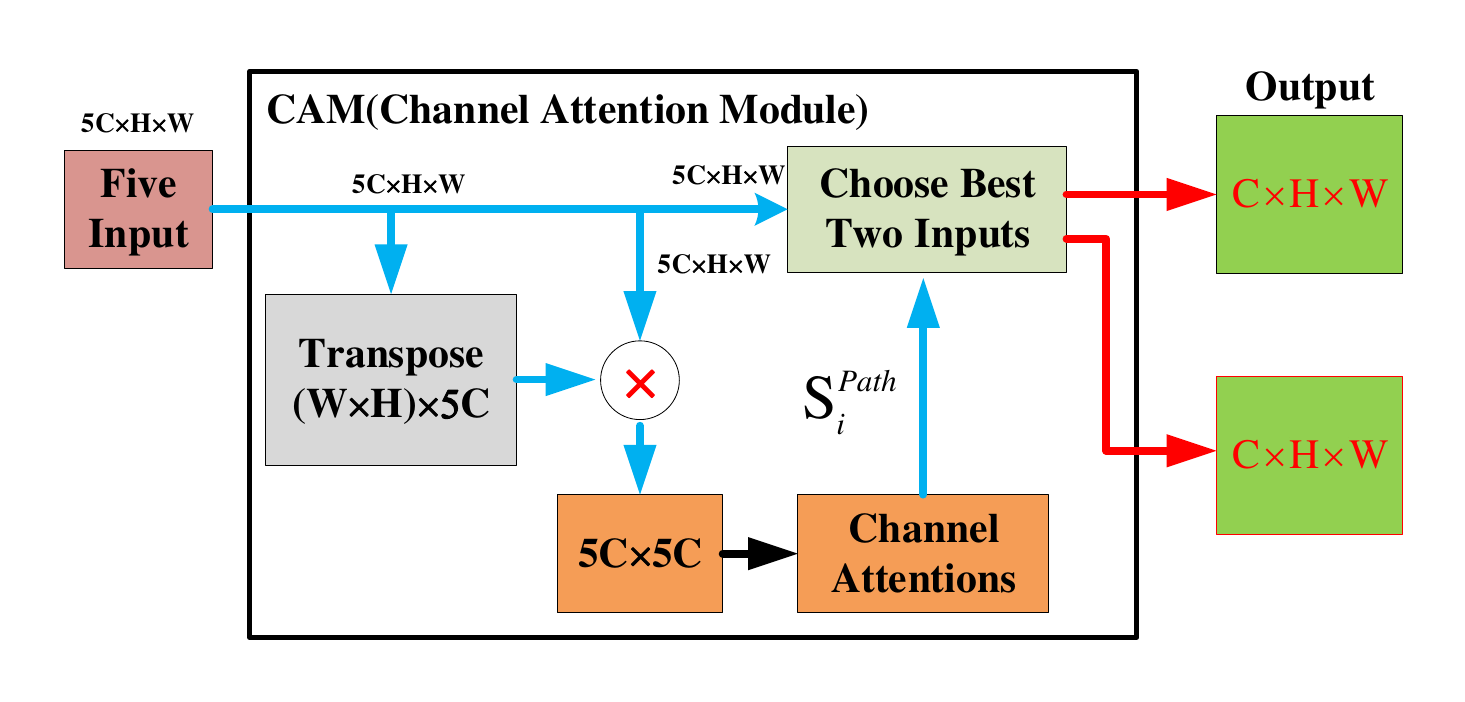}
  \vspace{-0.2cm}
} 
\caption{
Details of the proposed {\bf Channel Attention Module (CAM)}.
} 
\label{fig:CAM}
\vspace{-0.2cm}
\end{figure}

%%%%%%%%%%%%%%%%%%%%%%%%%%%%%%%%%%%%%%
%%%%%%%%%%%%%%%%%%%%%%%%%%%%%%%%%%%%%%
\subsection{Path Attention Module}
\label{sec:path:search}

To avoid the problem of combination exploration, we propose the \textbf{Path Attention Module (PAM)} to reduce the path search problem into a linear weight optimization. Our PAM considers the outputs from all previous layers and the current layer. It chooses the best two of them as inputs for the next cells to generate better segmentation results. 

Fig.~\ref{fig:PAM} depicts the details of this PAM.  Given a node, the idea of PAM aims at calculating the attention value of each path and choosing the number of outputs you want (denoted by red color in Fig.~\ref{fig:PAM}(a)) with higher attention to construct the final macrostructure. The PAM uses a $1\times1$ convolution and an interpolation operation to normalize all five inputs (denoted by four blue-arrows and one green-arrow in Fig.~\ref{fig:PAM}(b)) to the same size $C\times H\times W$. Then, they are sent to a {\bf Channel Attention Module (CAM)} to further calculate their channel attention.  Details of the CAM are depicted in Fig.~\ref{fig:CAM}.  For any pair of $i^{th}$ and $j^{th}$ inputs $F_{i}^c$ and $F_{j}^c$ on the $c^{th}$ channel, we perform an inner product between them to obtain channel attention $S^{c}_{i,j} \in R^{C \times C}$:
\vspace{-0.15cm}
\begin{equation}
 S_{i,j}^c = \frac{\exp(F_{i}^c\cdot F_{j}^c)}{\begin{matrix} \sum_{c=1}^C \exp(F_{i}^c\cdot F_{j}^c) \end{matrix}},
\vspace{-0.15cm}
\end{equation}
where $S_{i,j}^c$ is the impact of the $i^{th}$ path on the $j^{th}$ path on the $c^{th}$ channel. Then, the attention $S_i^{Path}$ of the $i^{th}$ path is obtained as follows:
\vspace{-0.2cm}
\begin{equation}
 S_i^{Path} = \sum_{j=1}^N \sum_{c=1}^C S_{i,j}^c,
\vspace{-0.15cm}
\end{equation}
where $N$ is the number of paths (or inputs).  Based on $ S_i^{Path}$, the best two are chosen from the $N$ paths.  As described above, our path selection approach is very different from AutoDeeplab~\cite{AutoDeepLab:CVPR2019} and HiNAS~\cite{HiNAS:CVPR2020} which take only three previous layers’ outputs as inputs.  Then, the Viterbi decoding algorithm is adopted to select only one input with the highest probability to generate the final result.  However, our method  takes all previous layers’ and current layer’s output as inputs and choose the best one layer's feature maps for the next cell (we searched) for further processing.  Since more layers and inputs are considered, our method fuses much more information to generate better segmentation results. 
%%%%%%%%%%%%%%%%%%%%%%%%%%%%%%%%%%%%%%

%%%%%%%%%%%%%%%%%%%%%%%%%%%%%%%%%%%%%%

\subsection{Recursive Stage Partial (RSP)  Architecture for Light-Weight Semantic Segmentation}
\label{sec:path:csp}

Neural Architecture Search(NAS) can successfully identify neural network architectures that outperform the hand-designed ones. However, such success greatly relies on costly computation resources. To reduce computations and maintain our accuracy, we modify the concept of Cross Partial Stage(CSP)~\cite{CSPNet:CVPRW2020} to recursively use only half of the
channels to pass through the cell we searched for. Fig.~\ref{fig:N:cell} shows that we construct our cell architecture that recursively uses only half of the channels to pass through the operation. Before entering the next cell, we concatenate the output and the unused part of the input. This RSP architecture can make the search process much more efficient and results in a light-weight design for semantic segmentation. 
%%%%%%%%%%%%%%%%%%%%%%%%%%%%%%%%%%%%%%

\begin{table*}[t]
\caption{
Performance evaluations of the model on Cityscapes validation set.
%\vspace{-0.2cm}
}
\centerline{
%\vspace{-0.3cm}
\setlength{\tabcolsep}{3mm}{
\begin{tabular}{lccccc}
\toprule
    Methods   & \multicolumn{1}{l}{Backbone}  & \multicolumn{1}{l}{Coarse} & \multicolumn{1}{l}{ImageNet} & \multicolumn{1}{l}{mIOU(\%)} & \multicolumn{1}{l}{Params(M)} \\\hline
    
    PSPNet~\cite{PSANet:ECCV2018} & Dilated-ResNet-101  & $\times$  & $\checkmark$  &76.2  & $\times$   \\
    PSANet~\cite{PSANet:ECCV2018} & Dilated-ResNet-101   & $\times$  & $\checkmark$  & 77.3  & $\times$ \\
    PADNet~\cite{PADNet:CVPR2018} & Dilated-ResNet-101   & $\times$  & $\checkmark$  & 78.1  & $\times$  \\
    DenseASPP~\cite{DenseASPP:CVPR2018}  & WDenseNet-161    & $\times$  & $\checkmark$  & 77.8  & $\times$ \\
    DeepLabv3~\cite{DeepLabv3:2017}  & ResNet-101   & $\checkmark$  & $\checkmark$  & 79.5  & $\times$  \\\hline
    
    Auto-DeepLab~\cite{AutoDeepLab:CVPR2019}  & $\times$  & $\checkmark$  & $\times$   & 80.3  & 44.42 \\
    HRNetV2~\cite{HRNet:arxiv2019} & $\times$  & $\times$ & $\checkmark$  & 80.9  & 69.06 \\\hline
    Ours  & $\times$  & $\times$  & $\checkmark$  &  81.4  & 68.67 \\
    Ours + RSP  & $\times$  & $\times$  & $\checkmark$  & 81.0  & 17.20 \\
    
\bottomrule
\end{tabular}}
%\vspace{-0.2cm}
}    
\label{tbl:experiment}
%\vspace{-0.4cm}
\end{table*}

%%%%%%%%%%%%%%%%%%%%%%%%%%%%%%%%%%%%%%

\begin{table*}[t]
\caption{
Performance evaluations of the model on Cityscapes validation set. Training with the Mapillary Vistas dataset.
\vspace{-0.2cm}
}
\centerline{
%\vspace{-0.3cm}
\setlength{\tabcolsep}{3mm}{
\begin{tabular}{lcccc}
\toprule
    Methods   & \multicolumn{1}{l}{Backbone}  & \multicolumn{1}{l}{Mapillary}  & \multicolumn{1}{l}{mIOU(\%)} & \multicolumn{1}{l}{Params(M)} \\\hline
    
    Mapillary~\cite{MapillaryVistas:ICCV2017} & ResNeXt-101 & $\checkmark$  & 80.6  & $\times$ \\
    HANet~\cite{HANet:CVPR2020} & ResNet-101     & $\checkmark$    & 81.7  & $\times$ \\ 
    HRNetV2+OCR~\cite{HRNet:arxiv2019} & HRNetV2  & $\checkmark$  & 81.8   & 70.37 \\
    DecoupleSegNets  & Wide-ResNet  & $\checkmark$   & 81.6  & $\times$ \\
    DCNAS~\cite{DCNAS:CVPR2021}  & $\times$  & $\checkmark$   & 81.3  & $\times$ \\\hline
    Ours  & $\times$  & $\times$    & 81.4  & 68.67 \\
    Ours  & $\times$  & $\checkmark$    & 82.1  & 68.67  \\
    Ours + RSP  & $\times$  & $\checkmark$    & 81.7  & 17.20 \\
    
\bottomrule
\end{tabular}}
%\vspace{-0.2cm}
}    
\label{tbl:experiment2}
%\vspace{-0.4cm}
\end{table*}

%%%%%%%%%%%%%%%%%%%%%%%%%%%%%%%%%%%%%

%------------------------------
\begin{figure}[t]
\centerline{
  \includegraphics[width=\linewidth]{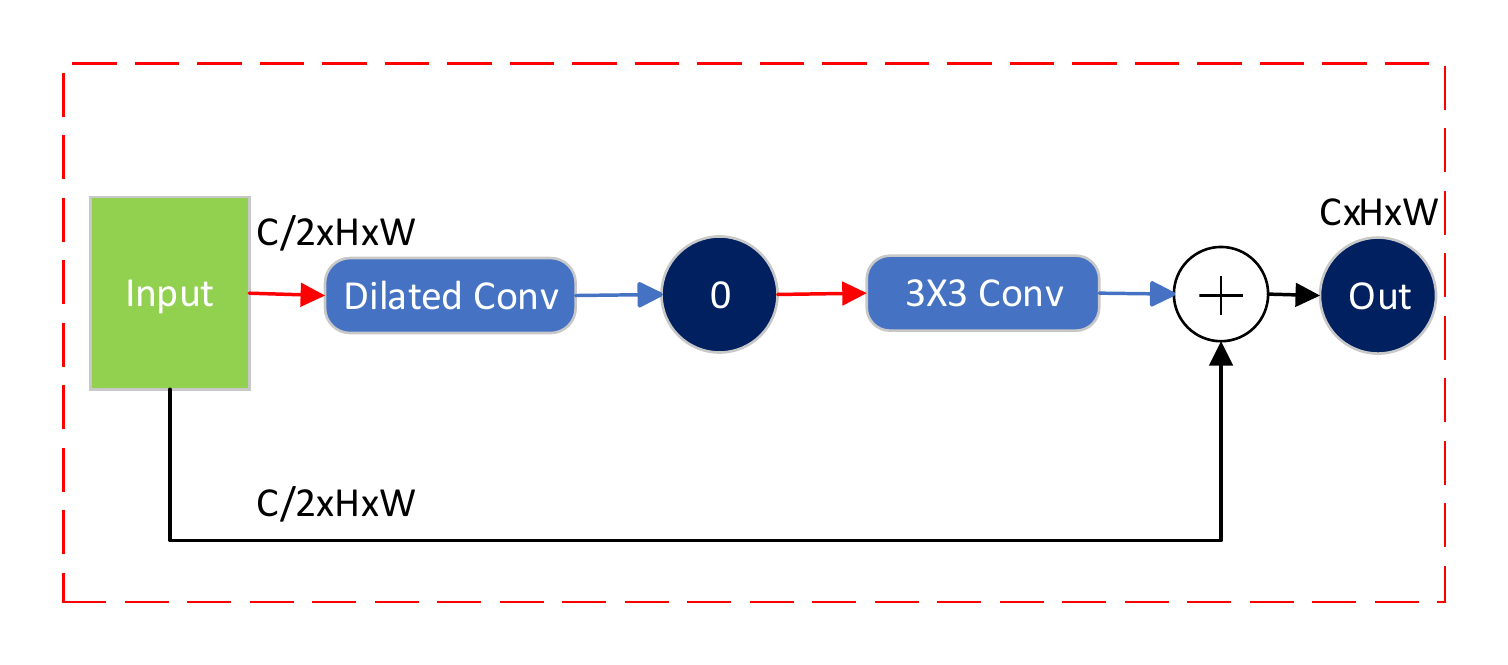}
  \vspace{-0.2cm}
}  
\caption{
DARTS searched architecture on the Cityscapes dataset.
\vspace{-0.2cm}
} 
\label{fig:N:cell_result}
\vspace{-0.2cm}
\end{figure}
%------------------------------

%%%%%%%%%%%%%%%%%%%%%%%%%%%%%%%%%%%%%%
%%%%%%%%%%%%%%%%%%%%%%%%%%%%%%%%%%%%%%

\section{Experimental Results}

Experiments are conducted in the Cityscapes~\cite{Cityscapes:CVPR2016} urban scene understanding dataset for evaluation.
Auto-DeepLab~\cite{AutoDeepLab:CVPR2019}, U-Net~\cite{UNet:MICCAI2015} and NasUnet~\cite{NAS:UNet:IEEEAccess2019} are compared in Cityscapes as baseline. We use the standard {\em mean intersection-over-union} (mIOU) as a performance evaluation metric for semantic segmentation.

%%%%%%%%%%%%%%%%%%%%%%%%%%%%%%%%%%%%%%
%%%%%%%%%%%%%%%%%%%%%%%%%%%%%%%%%%%%%%
\subsection{Results on Cityscapes scene segmentation}
\label{sec:evaluation}

The \textbf{Cityscapes} dataset~\cite{Cityscapes:CVPR2016} is a recent large-scale urban scene dataset containing a diverse set of stereo video sequences from 50 cities. Cityscapes dataset contains high quality pixel-level annotations of $5,000$ images with size $1,024 \times 2,048$. There are $2,975$, $500$, and $1,525$ for training, validation, and test images, respectively, and additional $20,000$ weakly annotated frames. It is an order of magnitude larger than similar previous datasets.

\begin{figure}[t]
\centerline{
  \includegraphics[width=0.95\linewidth]{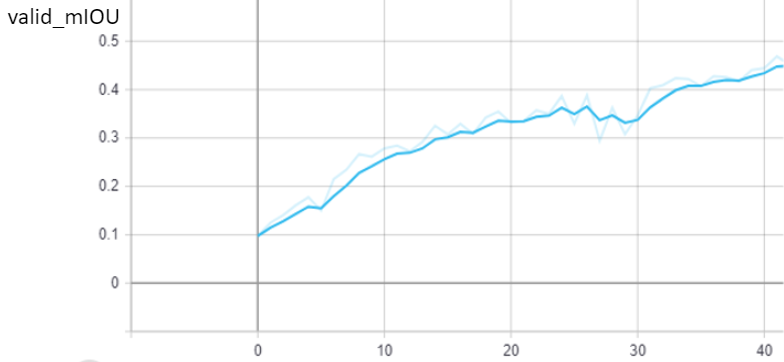}
  %\vspace{-0.2cm}
}  
\caption{
The validation accuracy (mIOU) during the 40 epochs of the macro structure search.
\vspace{-0.2cm}
} 
\label{fig:search_accuracy}
\vspace{-0.2cm}
\end{figure}

We consider 2 intermediate nodes in all cells with one input.
For each cell, we keep the channel numbers and the height and width of the feature tensor. Fig.~\ref{fig:N:cell_result} shows the searched cell architectures by NAS on the Cityscapes dataset. Fig.~\ref{fig:search_accuracy} shows that validation accuracy got more than 40\% during macro search. Compared to the accuracy in AutoDeepLab~\cite{AutoDeepLab:CVPR2019}, we got higher accuracy during search.
$512 \times 102$ random image crops are used. In DARTS search, batch size is 6 due to GPU memory limitation, architecture search optimization is conducted for $300$ epochs.
In the learning of network weight $w$, SGD optimizer with momentum $0.95$, and weight decay $0.0005$ are used. For learning the architecture, SGD optimizer with learning rate $0.005$ and weight decay $0.0001$ are used. The entire architecture search optimization takes about five days on two V100 GPUs.

\begin{figure*}[t]
\centerline{
  \includegraphics[width=0.95\linewidth]{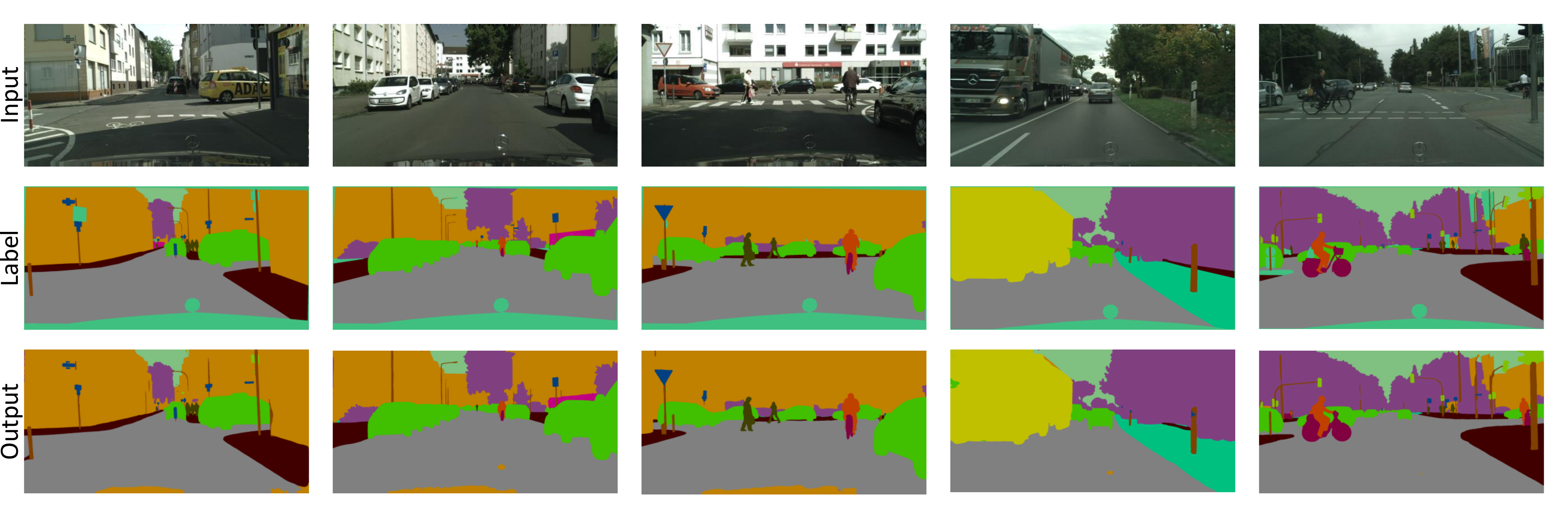}
  \vspace{-0.3cm}
}  
\centerline{
  \includegraphics[width=0.95\linewidth]{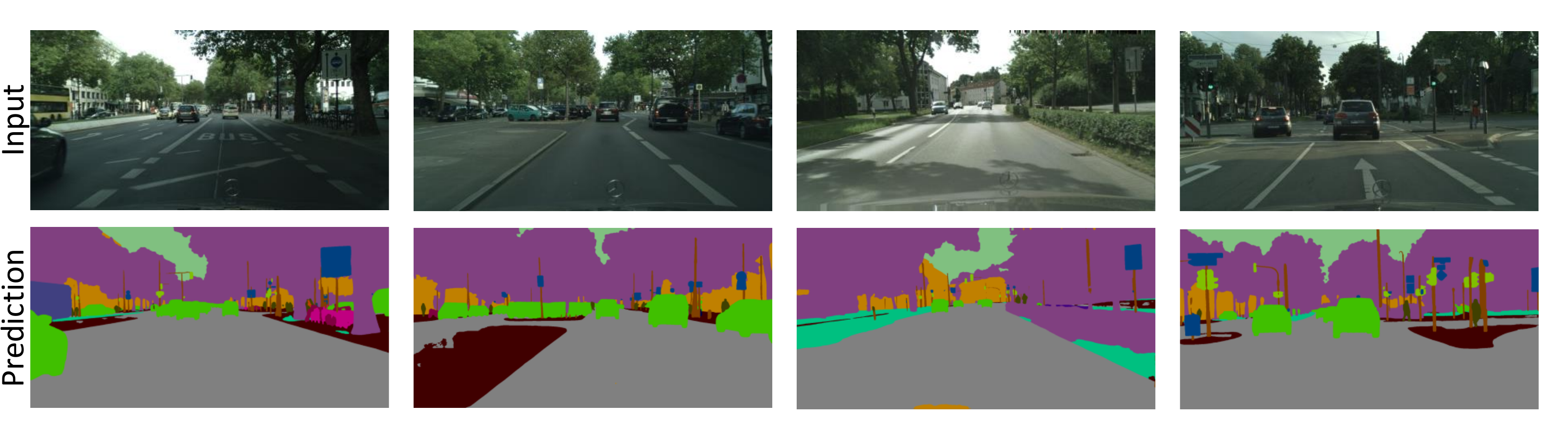}
  \vspace{-0.3cm}
}  
\caption{
Segmentation results of the proposed model on the Cityscapes (a) validation set and (b) test set.
} 
\label{fig:cityscapes_results}
\vspace{-0.4cm}
\end{figure*}

Table~\ref{tbl:experiment} shows that our NAS model outperforms the SoTA on the Cityscapes.
Without any pretraining, our best model significantly outperforms all the SoTA method. Last row of Table~\ref{tbl:experiment} shows that the light-weigh RSP design uses only 1/4 parameter size of SoTA methods but still outperforms them.  Fig.~\ref{fig:cityscapes_results} shows that the visualization of our model on Cityscapes~\cite{Cityscapes:CVPR2016} validation and test set.

The \textbf{Mapillary Vistas Dataset} is a large-scale street-level image dataset containing 25,000 high-resolution images annotated into 66/124 object categories of which 37/70 classes are instance-specific labels (v.1.2 and v2.0, respectively). Annotation is performed in a dense and fine-grained style by using polygons for delineating individual objects. Dataset contains images from all around the world, captured at various conditions regarding weather, season and daytime. Images come from different imaging devices (mobile phones, tablets, action cameras, professional capturing rigs) and differently experienced photographers. We also adopted the Mapillary Vistas Dataset~\cite{MapillaryVistas:ICCV2017} during our training procedure. Because of the class number of cityscapes is less than Mapillary Vistas Dataset, we have to map the category into the corresponding ones in Cityscapes. Table~\ref{tbl:experiment2} shows that our NAS model outperforms the SoTA on the Cityscapes with adopting Mapillary Vistas Dataset~\cite{MapillaryVistas:ICCV2017}.  Clearly, our light-weight RSP design still outperforms other SoTA methods even with only 1/4 parameter size. 

To better aggregate the context, we also adopted the Object-Contextual
Representations(OCR)~\cite{OCR:ECCV2020}. Table~\ref{tbl:experiment3} shows that our NAS model outperforms the other model with pretraining and Object-Contextual Representations(OCR)~\cite{OCR:ECCV2020} on the Cityscapes dataset.  Without using any backbone, our light-weight architecture still outperforms HRNET and DCNAS.  
  
%%%%%%%%%%%%%%%%%%%%%%%%%%%%%%%%%%%%%%

%%%%%%%%%%%%%%%%%%%%%%%%%%%%%%%%%%%%%%

\begin{table}[t]
\caption{
Performance evaluation on the Cityscapes validation set.
%\vspace{-0.2cm}
}
\centerline{
%\vspace{-0.3cm}
\setlength{\tabcolsep}{1mm}{
\begin{tabular}{lccc}
\toprule
    Methods   & \multicolumn{1}{l}{OCR}  & \multicolumn{1}{l}{ImageNet}  & \multicolumn{1}{l}{mIOU(\%)} \\\hline
    
    HRNetV2~\cite{HRNet:arxiv2019} & $\times$   & $\times$  & 76.16  \\
    HRNetV2   & $\checkmark$   & $\times$  & 78.2  \\
    HRNetV2   & $\times$   & $\checkmark$  & 80.9  \\
    HRNetV2   & $\checkmark$   & $\checkmark$  & 81.6  \\
    DCNAS~\cite{DCNAS:CVPR2021}    & $\times$      & $\times$   & 81.9 \\\hline
    
    Ours & $\times$  & $\checkmark$    & 81.4  \\
    Ours & $\checkmark$  & $\checkmark$    & 83.2 \\
    Ours + RSP & $\checkmark$  & $\checkmark$    & 82.5 \\
    
\bottomrule
\end{tabular}}
%\vspace{-0.2cm}
}    
\label{tbl:experiment3}
%\vspace{-0.4cm}
\end{table}

%%%%%%%%%%%%%%%%%%%%%%%%%%%%%%%%%%%%%

%--------------------------------------
%--------------------------------------

%------------------------------
%------------------------------

%%%%%%%%%%%%%%%%%%%%%%%%%%%%%%%%%%%%%%
\subsection{Discussions}

We discuss practical considerations in deploying our method. First, is the sensitivity of the number $N$ of cells that is stacked during training. We use a fixed $N=4$ in this paper. More search parameters are involved and more time will be consumed for larger $N$, however better image segmentation performance might be obtained.  
%This paper focuses on improving neural architecture search for image segmentation. 
Secondly, the potential negative impact of our work is on the environmental aspect, that automatic model search requires significant GPU computation. Again, the proposed micro search can be completed in two day of computation on modern GPUs.

%%%%%%%%%%%%%%%%%%%%%%%%%%%%%%%%%%%%%%
%%%%%%%%%%%%%%%%%%%%%%%%%%%%%%%%%%%%%%
\section{Conclusion}

This paper designs a two-stage architecture that allows us to search micro structure with less memory resource for image segmentation. The complexity reduction in search space and path selection makes the entire architecture search optimization very efficient.  It takes about 2 days on two V100 GPUs to find the desired micro architecture. RSPNet makes our search procedure much more efficient and leads to our light-weight design for outstanding
semantic segmentation.

{\bf Future work} includes the generalization and consideration of additional computation structures and cells as a basis that can be incorporated into the proposed framework. The model can be replaced with other simple structures for memory- or power-aware applications.  Another issue is the selection of the number $N$.  Current version sets the same $N$ for each module.  If  the number of each module is  different, the system performance will be further improved.  In addition, we envision works to be carried out straightforwardly on the NAS for other computer vision and NLP tasks.

%%%%%%%%%%%%%%%%%%%%%%%%%%%%%%%%%%%%%
%%%%%%%%%%%%%%%%%%%%%%%%%%%%%%%%%%%%%

\bibliographystyle{aaai}
\bibliography{NAS_HRNet}

\end{document}